\relax
\documentclass[letterpaper]{article} 
\usepackage{aaai20}  
\usepackage{times}  
\usepackage{helvet} 
\usepackage{courier}  
\usepackage[hyphens]{url}  
\usepackage{graphicx} 
\usepackage{graphicx}  
\usepackage{url}
\usepackage{amsmath}
\usepackage{algorithm}
\usepackage{algpseudocode}
\algrenewcommand\algorithmicrequire{\textbf{Input:}}
\algrenewcommand\algorithmicensure{\textbf{Initialize:}}
\newtheorem{theorem}{Theorem}

\newcommand{\citet}[1]{\citeauthor{#1} \shortcite{#1}}
\newcommand{\citep}{\cite}

\usepackage[position=top]{subfig}
\nocopyright
\title{Cost-Accuracy Aware Adaptive Labeling for Active Learning}

\author{Ruijiang Gao\\
University of Texas at Austin\\
ruijiang@utexas.edu
\And
Maytal Saar-tsechansky\\
University of Texas at Austin\\
maytal.saar-tsechansky@mccombs.utexas.edu\\
}
 
\begin{document}

\maketitle

\begin{abstract}
Conventional active learning algorithms assume a single labeler that produces noiseless label at a given, fixed cost, and aim to achieve the best generalization performance for given classifier under a budget constraint. However, in many real settings, different labelers have different labeling costs and can yield different labeling accuracies. Moreover, a given labeler may exhibit different labeling accuracies for different instances. This setting can be referred to as active learning with diverse labelers with varying costs and accuracies, and it arises in many important real settings. It is therefore beneficial to understand how to effectively trade-off between labeling accuracy for different instances, labeling costs, as well as the informativeness of training instances, so as to achieve the best generalization performance at the lowest labeling cost. In this paper, we propose a new algorithm for selecting instances, labelers (and their corresponding costs and labeling accuracies), that  employs generalization bound of learning with label noise to select informative instances and  labelers so as to achieve higher generalization accuracy at a lower cost. Our proposed algorithm demonstrates state-of-the-art performance on five UCI and a real crowdsourcing dataset.
\end{abstract}

\section{Introduction}
Supervised learning has achieved great successes over the years and has a significant impact on practice in a growing variety of predictive tasks. In many settings, however, labels for training instances are not readily available, but can be acquired from different labelers at different costs; often, different labelers may exhibit varying labeling accuracies, and a given labeler can have different labeling accuracies across different instances, possibly as a result of experience or prior knowledge. Given this setting and a model induction algorithm, it is important to understand how to best select labelers and instances they will label so as to induce a model with the highest generalization performance for a given labeling cost. In practice, such challenges arise in important applications, where scientists, medical professionals, or crowd of lay workers can be used to label a possibly large number of instances. In recent years, large-scale  crowdsourcing and online labor market platforms, such as  Amazon  Mechanical  Turk (AMT),  have emerged to offer unprecedented scalability towards such tasks. Nevertheless, in the settings we consider here, selecting labelers' (and corresponding costs and accuracies) and the instances they will label to produce the best model at a given cost remains an open problem. 

Traditional active learning ~\cite{lewis1994sequential,gal2017deep,tang2019self} has received significant attention, and considers the problem of selecting instances for labeling when a single labeler produces labels at the same, fixed cost and with perfect labeling accuracy. Because labels of all instances are assumed to have the same accuracy and cost, traditional active learning frameworks aim to identify the most informative training instances from which to induce a model. However, in many real settings, acquiring labels from labelers presents greater complexity. AMT, for example, offers access to workers from around the world, with different expertise and varying costs. Indeed, prior social science research has shown that different payments lead to  different qualities of work, and that different relationships between payment and quality can arise at different times or for different tasks ~\cite{mason2009financial,kazai2011search,kazai2013analysis}. 

More recent work considered multiple noisy workers, yet assumed that either all labelers exhibit the same quality ~\cite{ipeirotis2014repeated,lin2014re,donmez2010probabilistic}, or that all labelers have the same cost per label and that the labeling quality is independent of the instance being labeled ~\cite{donmez2009efficiently,yan2011active,yan2014learning}. The closest work to the problem we consider here is by ~\citet{huang2017cost}, where  instance difficulty, labeler expertise, and varying costs across labelers are considered. ~\citet{huang2017cost} use a different criterion to select labelers and instance than the one we develop here; as we discuss below,  in the most common setting in practice, where labelers are not adversarial ~\cite{ipeirotis2014repeated,yan2011active,lin2014re,donmez2009efficiently}, this criterion appears to be prone to consistently choose low payment options, even when the labeling quality is poor and such choices undermine learning significantly. 

In this paper, we propose a novel criterion utilizing generalization bound of learning with label noise to evaluate the cost-effectiveness of labeler-instance pair. The criterion we propose is motivated by the goal of directly minimizing the generation error of the classifier. To the best of our knowledge, this work is the first to use generalization bound for guidance of selecting labeler-instance pairs in this setting. We empirically evaluate the effectiveness and robustness of our method for settings with different cost-accuracy trade-offs reported in prior work to arise in crowdsourcing markets, and for five UCI datasets and a real crowdsourcing dataset. Our results show that our approach offers state-of-the-art performance across settings.

\section{Related Work}

Active learning has been studied extensively and is getting more important with the emergence of modern artificial intelligence. Most active learning research has considered settings where there is only one perfect labeler. Active learning algorithms for these settings thus aim to select the most informative training instances to label, so as to reduce the number of instances. However, when crowdsourcing platforms are used to acquire labels, annotation is done by multiple noisy annotators, whose labels can be acquired at different costs and who may exhibit different levels of accuracies. Recently, there is some more research on learning from noisy labelers. ~\citet{donmez2009efficiently} and ~\citet{zheng2010active} estimate the accuracy rates of labelers and then selects for annotation labelers with high accuracies.   ~\citet{zhao2011incremental} actively select instances for labeling, but do not select labelers. All these works assume that a labeler exhibits the same accuracy for all instances they label. Yet, in practice, different workers may have different expertise or prior experience and can consequently exhibit different accuracies when labeling different instances. ~\citet{yan2011active} propose a probabilistic framework to estimate workers' accuracies and select the worker estimated to be the most accurate for a given instance. ~\citet{fang2014active} and ~\citet{ambati2010active} consider the different expertise of workers and aim to match instances with different worker with varying accuracies in the task domain; yet, these works do not consider varying labeling costs may be incurred by different labelers. ~\citet{geva2019more} consider labelers with varying accuracies and costs using estimated error reduction but do not consider selecting instances for labeling. The most closely related work is by~\citet{huang2017cost}: the proposed method estimates workers' labeling accuracies based on a small set of ground-truth data, and then estimates the value of acquiring the label for a given instance from a given worker as the weighted labeling accuracy divide by worker's cost. As we discuss in more detail below and reflected in our empirical evaluations, in the most common setting in practice, when labelers are not adversarial~\cite{ipeirotis2014repeated,yan2014learning}, this heuristic criterion is prone to select low payments.

Some prior work considered the cost-effectiveness of majority voting by multiple labelers for same instance as compared to singly-labeled data. These works considered workers who exhibit the same accuracy and cost.  ~\citet{ipeirotis2014repeated} shows how in some cases majority voting can improve the performance of given classifier for a given cost, and ~\citet{lin2014re} demonstrates how the optimal choice depends on the dataset, classifier, and labeling accuracy. Yet, these works do not address how to effectively trade-off performance and cost. Importantly for this work, the trade-off between acquiring a single or multiple labels per instance can also be viewed as a special case of having multiple labelers of varying costs and accuracies. Hence,  a method that can effectively select amongst different labelers of accuracies can also apply to select whether or not to acquire multiple labels for a given instance.
Other research explored the generalization error bounds for learning with label noise. ~\citet{simon1996general} and ~\citet{aslam1996sample} study the error bounds for learning from noisy labels for PAC-learnable concepts. ~\citet{kearns1998efficient} develop a bound for concepts that are Statistical-Query-learnable. We rely on these theoretical results and propose a novel criterion to select instance-labeler pairs that minimize the generalization error 
and achieve state-of-the-art results.

\section{Problem Statement}
Suppose we have a dataset $D = \{x_i,y_i\}_{i=1}^N$, concept $\mathcal{C}$, an unlabeled set $\mathcal{U} = \{x_i\}_{n_l + 1}^{N}$, and a set of labelers $\mathcal{A} = \{a_1, \dots , a_n\}$ , who exhibit costs $\{c_1,\dots,c_n\}$ and label accuracies $\{\rho_1,\dots,\rho_n\}$, respectively. In addition, suppose that an initial labeled dataset is available with ground-truth labels $\mathcal{L} = \{x_i,y_i\}_{1}^{n_l}$  that have also been labeled by each of the labelers in $\mathcal{A}$. We assume the most common settings where labelers are not adversarial, i.e. their labeling accuracy rates are higher than 50\%~\cite{ipeirotis2014repeated,yan2011active,lin2014re,donmez2009efficiently}.

\begin{figure}
    \centering
    \includegraphics[width=3.3in]{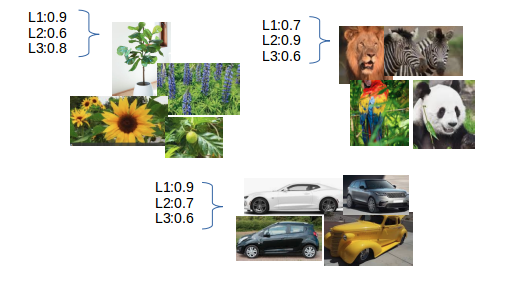}
    \caption{Diverse labelers' performance on different image categories: Flora, Fauna and Cars. Labels L1, L2, L3 demonstrate different labeling accuracies on different category, and their overall accuracies are 0.83, 0.73 and 0.66, respectively (assuming equal numbers of instances in each category).}
    \label{fig:prob_stat}
\end{figure}

Further, labelers can have different expertise for different instances.  Thus, for example, in an image classification task, Amy may be an expert at identifying species of flora while Bob excels in identifying fauna. We illustrate this notion in Figure \ref{fig:prob_stat}, where labelers, $L1$, $L2$, and $L3$, have diverse expertise across different image categories. If each image category has the same number of samples, the overall labeling accuracies for $L1$, $L2$, $L3$ are 0.83, 0.73, 0.66 respectively. Yet, each labeler exhibits higher (and lower) labeling accuracies on some of the categories. Empirical work  ~\cite{kazai2011search,kazai2013analysis,mason2009financial} show that labelers with higher labeling accuracies often incur higher costs. We follow the setting considered in prior work, and for simplicity assume the price levels of 3, 2, and 1, for labelers $L1$, $L2$, and $L3$, respectively. 

Finally, we consider an iterative setting, where at each iteration, a labeler from $\mathcal{A}$ is selected for labeling a selected instance from $\mathcal{U}$. Given a limited budget $B$, we aim to acquire labels from labelers for certain instances so as to induce a classifier with the best generalization performance. Thus, an algorithm for selecting labelers and instances ought to decide from which labelers and for what instances to acquire labels so as to yield the best generalization performance with a given budget; for example, it ought to determine whether it would be more cost-effective to acquire flora labels from the more accurate (and pricier) L1 than acquiring lower accuracy labels for 3 different flora images from L3, thereby compiling a larger training set for learning at the same cost.

\section{Algorithm}
Recall that our approach aims to select instance-labeler pairs so as to improve the generalization performance for a given labeling budget.  Below, we first discuss briefly how we quantify instance usefulness. Because the closest work to the contributions we present here is by~\citet{huang2017cost}, we then outline the main elements of the CEAL algorithm \cite{huang2017cost} and then describe how our algorithm builds on this contributions. 

\subsection{Instance Usefulness}
There are many algorithms that propose various criteria for selecting instances for labeling; the more prominent measures include: uncertainty sampling that uses the posterior
probability of predicted class~\cite{gal2017deep,lewis1994sequential}, model expected change that selects the sample the affects the model most~\cite{freytag2014selecting} and data diversity that chooses the data that helps labeled pool better represent the underlying population~\cite{sener2017active,nguyen2004active}. In general, any measure for quantifying  instance usefulness can be used in our approach. In this paper, our main focus is on a cost-effective selection of labelers, we simply follow the setting in ~\cite{huang2017cost} and use uncertainty sampling as shown in Equation \ref{eqn:uncertainty}. $P(y|x_j)$ is the posterior probability predicted by the classifier trained at current iteration.


\begin{equation}
    \label{eqn:uncertainty}
    r(x_j) = 1 - \max_{y\in \mathcal{Y}} P(y|x_j)
\end{equation}

\subsection{CEAL ~\cite{huang2017cost}}
CEAL estimates annotators' labeling accuracies of given instance $x_j \in \mathcal{U}$ based on the accuracy of labelers' respective responses on the labeled set $\mathcal{L}$ with ground truth labels. Specifically, the accuracy of labeler $i$ for instance $x_j$, $\rho_i(x_j)$ is a weighted mean of labeling accuracy, weighted by the similarity between $x_j$ and each of its nearest neighbors $x_k \in \mathcal{N}(x_j)$, as shown in Equation \ref{eqn:accuracynn} where $0\leq s(x_k, x_j) \leq 1$, $\mathcal{N}(x)$ represents that nearest neighbors of $x$ in $\mathcal{L}$ and $\sum_k s(x_k, x_j) = 1$.
\begin{equation}
    \label{eqn:accuracynn}
    \rho_i(x_j) = \sum_{x_k \in \mathcal{N}(x_j)} s(x_k, x_j)I[y_k == \hat{y}_{ik}]
\end{equation}
The final instance-labeler pair is selected by Equation \ref{eqn:ceal} and \ref{eqn:selceal}. At each iteration, the product $q_i(x_j)r(x_j)$ is computed for every instance-labeler pair $(x_j,a_i), x_j\in \mathcal{U}, a_i \in \mathcal{A}$ pair, and the pair with maximum product is selected for labeling. As a result, CEAL will select samples that are quite useful and have labelers that are good enough but cheap.

\begin{equation}
    \label{eqn:ceal}
    q_i(x_j) = \frac{\rho_i(x_j)}{c_i}
\end{equation}

\begin{equation}
    \label{eqn:selceal}
    (x^\star, a^\star) = \underset{a_i, x_j}{\arg\max} q_i(x_j)r(x_j)
\end{equation}

However, this heuristic tends to select instance-labeler pairs that maximize \textit{labeling} accuracy per cost, it does not assess the implications of different labeling accuracies and costs on generalization performance. To illustrate how CEAL prioritizes amongst different labelers, suppose we have five labelers whose labeling costs are 1,2,3,4,5 respectively. We further suppose labelers are not adversarial~\cite{ipeirotis2014repeated,yan2011active,lin2014re,donmez2009efficiently}, thus $\rho_i(x_j)$ is greater than 0.5 for all $i$. When a low cost labeler offers near-random accuracy, while all others offer perfect accuracy, it is easy to see that CEAL will always prefer the least costly labeler with near-random labeling accuracy. 
\begin{equation}
    \frac{\rho_i(x_j)}{1} \geq  \frac{0.5}{1}\geq\frac{\max \rho(x)}{2} = \frac{1.0}{2} > \frac{1.0}{3} > \frac{1.0}{4} > \frac{1.0}{5}
\end{equation} 
Similarly, in the example shown in Figure \ref{fig:prob_stat}, given all labelers' accuracies are above 0.5, CEAL will select the labeler with the lowest cost.

\subsection{Proposed Algorithm}


As we discussed above, in CEAL, the value of a labeler is quantified by the ratio between the labeler's labeling quality and cost, and as such does not necessarily reflect the impact on generalization performance. We seek to develop an algorithm that aims to address this to identify labeler-instance pairs that would have the greatest benefit to generalization performance per cost. 


However, generalization error is intractable in most supervised learning problems. There are some prior works that explore different ways to estimate it. In the context of traditional active learning,
~\cite{roy2001toward} proposes to use (empirical) Estimated Error Reduction (EER) as an estimation to the generalization error reduction to select useful training instances for labeling. 
~\cite{settles2008multiple,freytag2013labeling} maximize Expected Model Change (EMC) by evaluating expected changes in model parameters, but this approach lacks a theoretical connection to error reduction.  
~\cite{freytag2014selecting} proposes to use expected model output change as an upper bound to generalization error. While it gives no guarantee when we are interested in maximization, it shows good performance in active learning problems. 
However, all these approaches consider settings with a single and perfectly accurate labeler, and it is unclear how they can apply in our setting when multiple noisy workers are present.

Meanwhile, research on generalization error bound for different concepts when learning from noisy labels, provides an upper bound on the decreasing speed on the generalization error of concept classes of interest ~\cite{simon1996general,aslam1996sample,kearns1998efficient}, but is rarely used in empirical research. The upper bound of the generalization error can also be thought as incorporating uncertainty of the classification error. Inspired by the notion of guiding acquisition by expected error reduction, we propose a novel criterion that utilizes the theoretical results on generalization bound for learning with noisy labels to select cost-effective labelers. Based on the generalization bound proposed in Theorem \ref{thm:gbound}~\cite{simon1996general,aslam1996sample}, our algorithm combines theoretical analysis into active learning to select the labeler that minimize error as shown in Equation \ref{eqn:selcriterion}, where $\hat{\rho_i}$, and $n_i$ denote the estimated label accuracies and number of samples labeler $i$ can purchase under a fixed budget. Noted VC dimension of a classifier measures the size of the largest finite subset of X that it is capable of classifying correctly (shatter). A higher VC dimension corresponds to weaker inductive bias. For simplicity, we treat the fail probability term as a constant since VC dimension is not known. 

\begin{theorem} ~\cite{simon1996general,aslam1996sample}
\label{thm:gbound}
PAC learning a function class $\mathcal{F}$ with Vapnik-Chervonenkis dimension $\text{VC}(\mathcal{F})$ in the presence of classification noise $\rho$ and fail probability $\delta$ requires a sample of size
\begin{equation}
    \label{eqn:gbound}
    \Omega\Big(\frac{\text{VC}(\mathcal{F})}{\epsilon(2\rho - 1)^2} + \frac{\log(1/\delta)}{\epsilon(2\rho - 1)^2}\Big)
\end{equation}
\end{theorem}

\begin{align}
    i^\star & = \underset{i}{\arg\min} \frac{1}{(2\hat{\rho_i} - 1)\sqrt{n_i}}\\
    \label{eqn:selcriterion}
    & = \underset{i}{\arg\max} (2\hat{\rho_i} - 1)\sqrt{n_i}
\end{align}
The cost-normalized benefit to generalization performance from selecting labeler $i$ to label instance $x_j$ can thus be captured by the term in  Equation \ref{eqn:expm}. 
\begin{equation}
    \label{eqn:expm}
    q_i(x_j) = \frac{2\rho_i(x_j) - 1}{\sqrt{c_i}}
\end{equation}
However, recall that we aim to select labelers that lead to lowest generalization error. We should therefore consider the expected \textit{cumulative} accuracies for all options so far. Also, importantly, ~\cite{lin2014re} shows how more accurate labels may be preferred early on the learning curve, while cheaper and noisier labels may be more cost-effective for learning when number of samples are sufficiently large. Methods such as CEAL and Equation \ref{eqn:expm} neglect this; hence, we propose a second, adaptive criterion shown in Equation \ref{eqn:curem}, where $\rho_0,n_0$ is the estimated accuracy and number of instances so far, and $b$ denotes the unit budget we consider for estimating the future expected generalization error. 

\begin{equation}
    \label{eqn:curem}
    q_i(x_j) = (2\frac{\rho_0 n_0 + \rho_i(x_j)\lfloor\frac{b}{c_i}\rfloor}{n_0 + \lfloor\frac{b}{c_i}\rfloor} - 1)\sqrt{n_0 + \lfloor\frac{b}{c_i}\rfloor}
\end{equation}

\begin{equation}
    \label{eqn:efffectiveness}
    E(a_i, x_j) = r(x_j)q_i(x_j)
\end{equation}

\begin{equation}
    \label{eqn:choice}
    (x^\star, a^\star) = \underset{a_i, x_j}{\arg\max} E(a_i, x_j) 
\end{equation}

Similar to CEAL, we estimate labelers' accuracies using Equation \ref{eqn:accuracynn}. At each iteration, we calculate Equation \ref{eqn:efffectiveness} for each instance-labeler pair and select the pair with the highest value, as in Equation \ref{eqn:choice}; after the label from the chosen labeler for the instance is acquired, the labeled instance is added to current training set, and next iteration begins. The acquisitions continue until the budget is exhausted. The complete Generalization Bound based Active Learning (GBAL) algorithm based on the criterion in Equation \ref{eqn:expm}, and for Adaptive GBAL (AGB)\footnote{Code is availble in \url{https://github.com/ruijiang81/AGB}} are shown in Algorithms \ref{alg:gceal}, \ref{alg:gbal}.

\begin{algorithm}[h]
    \caption{Generalization Bound based Active Learning \textbf{(GBAL)}}
    \label{alg:gceal}
    \begin{algorithmic}
    \Require \\
    $L$: a small labeled set \\
    $U$: the pool of unlabeled data for active selection \\ 
    $A$: all possible labelers \\
    $\hat{Y}$:  the labels given by all labelers in $A$ on $L$
    \Repeat 
    \For { each $x_j \in U$ and labeler $a_i$} 
        \State calculate the uncertainty for $x_j$ in Equation \ref{eqn:uncertainty}
        \State calculate expected generalization bound as in Equation \ref{eqn:expm} or \ref{eqn:curem}
        \State calculate the effectiveness as in Equation \ref{eqn:efffectiveness}
    \EndFor
    \State Select the pair $(x^\star, a^\star)$ in Equation \ref{eqn:choice}.
    \State Query the label of $x^\star$ from $a^\star$, denoted by $\hat{y}^\star$. 
    \State $L = L \cup (x^\star, \hat{y}^\star)$; $U = U \setminus x^\star$. 
    \State Train classifier on $L$ and test it on test set.
    \Until {the budget is used up}
    \end{algorithmic}
\end{algorithm}

\begin{algorithm}[h]
    \caption{Adaptive GBAL \textbf{(AGB)}}
    \label{alg:gbal}
    \begin{algorithmic}
    \Require \\
    $L$: a small labeled set \\
    $U$: the pool of unlabeled data for active selection \\ 
    $A$: all possible labelers \\
    $\hat{Y}$:  the labels given by all labelers in $A$ on $L$ \\
    $b$:   unit budget for estimating Equation \ref{eqn:curem}
    \Ensure \\
    $\rho = 1$ 
    \Repeat 
    \For { each $x_j \in U$ and labeler $a_i$} 
        \State calculate the uncertainty for $x_j$ in Equation \ref{eqn:uncertainty}
        \State calculate the the expected generalization bound as in Equation \ref{eqn:curem} 
        \State calculate the effectiveness as in Equation \ref{eqn:efffectiveness}
    \EndFor
    \State Select the pair $(x^\star, a^\star)$ in Equation \ref{eqn:choice}.
    \State $\rho = (\rho n + \hat{\rho}_\star(x_j))/(n + 1)$
    \State Query the label of $x^\star$ from $a^\star$, denoted by $\hat{y}^\star$. 
    \State $L = L \cup (x^\star, \hat{y}^\star)$; $U = U \setminus x^\star$. 
    \State Train classifier on $L$ and test it on test set.
    \Until {the budget is used up}
    \end{algorithmic}
\end{algorithm}

\section{Experiment}
We compare our methods with four baselines:
\begin{itemize}
    \item \textbf{ALC:} ~\cite{yan2011active} ALC selects the most uncertain sample from the unlabeled set and use the most accurate labeler for annotation at each iteration.
    \item \textbf{CEAL
    :} 
    ~\cite{huang2017cost} CEAL selects instance-labeler pair that maximize Equation \ref{eqn:ceal}.
    \item \textbf{All:} Select the most uncertain sample and use majority voting based on all labelers  annotations for the sample at each iteration.
    \item \textbf{Random:} Select the most uncertain sample and randomly choose a labeler from the set of all labelers to annotate the sample at each iteration.
\end{itemize}
The main goal of the evaluation is to compare the effectiveness of labelers (costs and accuracies) chosen by different algorithms. We therefore do not consider methods that randomly select a sample from unlabeled set. The effectiveness of uncertainty sampling has been established in prior work~\cite{huang2017cost,lewis1994heterogeneous,gal2017deep}. Algorithm \ref{alg:gceal} and \ref{alg:gbal} are referred to as \textbf{GB} and \textbf{AGB} in this section. 

\subsection{Label Simulation}
We use the publicly available UCI datasets, and therefore we simulate the labels produced by different labelers for these datasets. The label generating process we use is similar to that in ~\cite{yan2011active}. Specifically, in order to create diverse labelers, we first create 30 clusters using KMeans~\cite{jain1999data} for each dataset. In addition, as in ~\cite{huang2017cost}, we simulate five labelers with cost levels: 5, 4, 3, 2, 1  which are associated with overall labeling accuracies from high to low, respectively. Each labeler is an `expert'' in some random set of clusters by exhibiting a high probability of correctly labeling instances from the corresponding cluster. In particular, the probabilities that a labeler  correctly labels instances in her ``expert'' clusters are 0.95, 0.925, 0.9, 0.875, and 0.85; these probabilities for ``non-expert'' clusters are 0.61, 0.585, 0.56, 0.535, and 0.51 respectively. The resulting overall worker accuracies on UCI datasets are shown in Table \ref{tab:uciworker}. This process produces a diverse label distribution, where different labelers also incur different costs. As demonstrated in Table \ref{tab:uciworker}, given the KMeans produce different clusters for different datasets, a labeler of a given cost can yield different overall labeling accuracy across different datasets -- this allows us to explore the robustness of our proposed algorithm under a wide variety of price-accuracy trade-offs.

\begin{table}
    \centering
    \begin{tabular}{c|c|c|c|c|c}
    \hline
     Labeler&  W1 & W2 & W3 & W4 & W5 \\\hline
     Pen Digits & 0.90 & 0.79 & 0.70 & 0.62 & 0.56 \\\hline
     Audit & 0.94 & 0.91 & 0.71 & 0.67 & 0.66 \\ \hline
     Mushroom & 0.92& 0.82& 0.76& 0.68& 0.61\\ \hline
     Spambase & 0.95 & 0.81 & 0.79 & 0.71 & 0.58 \\ \hline
     German & 0.93 & 0.87 & 0.74 & 0.68 & 0.57 \\ \hline
    \end{tabular}
    \caption{Worker Label Accuracy on UCI Datasets}
    \label{tab:uciworker}
\end{table}

\subsection{UCI Dataset}

\begin{table*}[h]
    \centering
    \begin{tabular}{c|c|c|c|c|c}
        Methods  &  Audit & Pen Digits & Spambase & German& Mushroom\\ \hline
        CEAL& 1.04& 1.01 & 1.01 & 1.03&1.01\\\hline 
        Random& 3.01& 3.01& 3.04& 3.01&3.03\\\hline
        ALC& 4.53& 4.63&4.59& 4.54&4.40\\\hline
        GB&2.97& 3.14& 1.50& 2.68&2.30\\\hline
        AGB&3.26&4.12& 2.57& 3.59&3.67\\\hline
    \end{tabular}
    \caption{Average cost on UCI Datasets of different methods}
    \label{tab:uci_cost}
    \centering
    \begin{tabular}{c|c|c|c|c|c}
        Methods  &  Audit & Pen Digits & Spambase& German& Mushroom\\ \hline
        CEAL& 0.67& 0.51 & 0.67 & 0.59&0.57\\\hline 
        Random& 0.81& 0.70& 0.76& 0.74&0.71\\\hline
        ALC& 0.95& 0.94& 0.93 & 0.92&0.79\\\hline
        GB&0.91& 0.82& 0.81& 0.82&0.73\\\hline
        AGB&0.93&0.88&0.88 & 0.85&0.77\\\hline
    \end{tabular}
    \caption{Average Label Accuracy on UCI Datasets of different methods}
    \label{tab:uci_la}
    \centering
    \begin{tabular}{c|c|c|c|c|c}
        Methods  &  Audit & Pen Digits & Spambase & German& Mushroom\\ \hline
        CEAL& 194.8& 200 & 200 &  195.4 & 200\\\hline 
        Random& 68.3& 68.2& 67.8 & 68.1 & 67.4\\\hline
        ALC& 46.0& 44.7&45.3& 45.6&47.3\\\hline
        GB&69.4& 66.4& 136.7& 77.2&91.8\\\hline
        AGB&63.4&50.2& 81.5 & 57.5&57.4\\\hline
    \end{tabular}
    \caption{Average Number of Queries on UCI Datasets of different methods}
    \label{tab:uci_query}
\end{table*}

We evaluated our approach using the following datasets: \textit{German, Mushroom, Pen Digits, Spambase, Audit} from UCI Machine Learning Repository ~\cite{bache2013uci}. The statistics of these five datasets can be found in Table \ref{tab:uci}.

\begin{table}
    \centering
    \begin{tabular}{c|c|c}
         & \#Instance & \#Feature \\\hline
        German & 1000& 24\\\hline
        Mushroom & 8124& 117\\\hline
        Pen Digits & 10992& 16\\\hline
        Spambase& 4601& 57\\\hline
        Audit & 776& 24\\\hline
    \end{tabular}
    \caption{Statistics of five UCI Datasets}
    \label{tab:uci}
\end{table}

\begin{figure*}
\centering
\begin{tabular}{cccc}

\includegraphics[width=.3\textwidth]{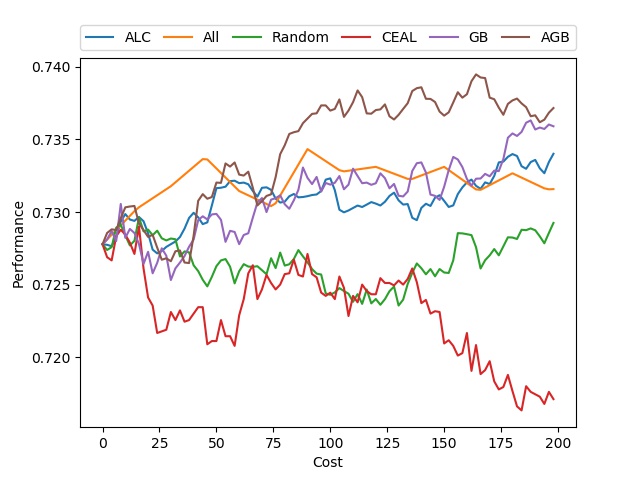} &
\includegraphics[width=.3\textwidth]{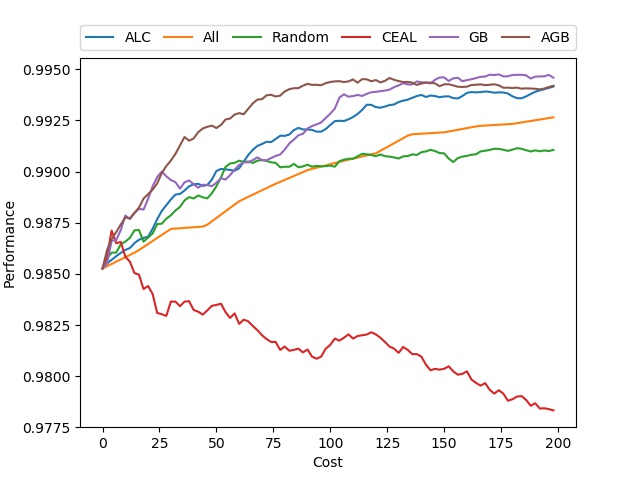} &
\includegraphics[width=.3\textwidth]{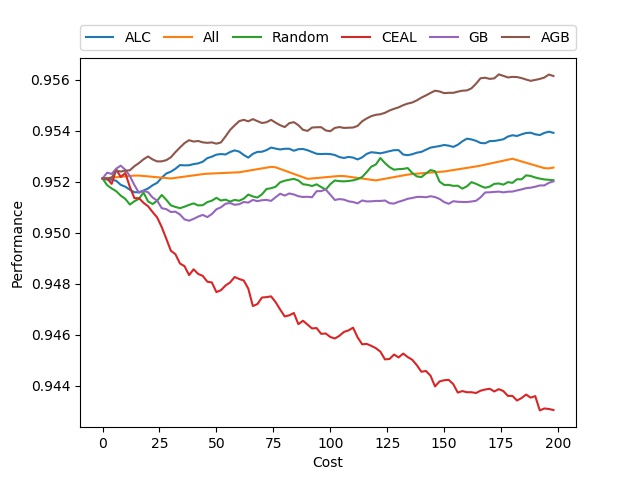} \\
\text{(a) German}  & \text{(b) Mushroom} & \text{(c) Pen Digits}  \\[6pt]
\end{tabular}
\begin{tabular}{cccc}

\includegraphics[width=.3\textwidth]{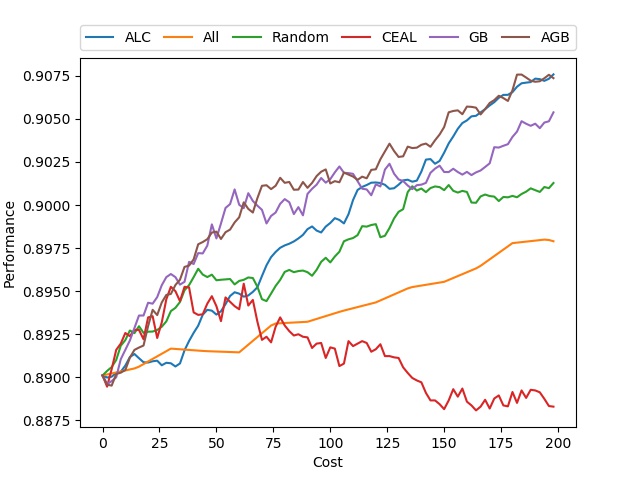} &
\includegraphics[width=.3\textwidth]{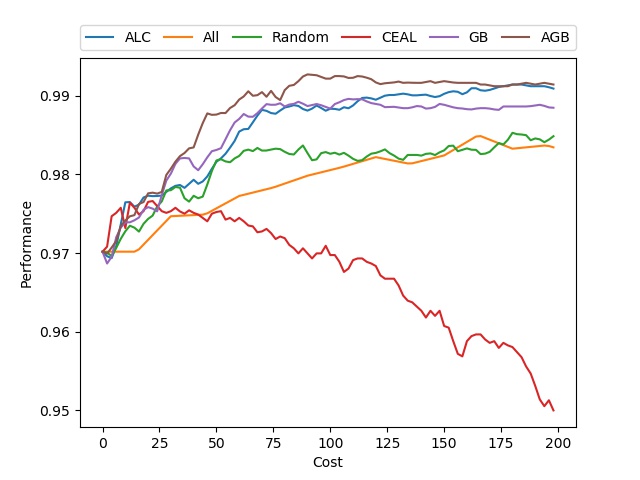} \\
\text{(d) Spambase}  & \text{(e) Audit}  \\[6pt]
\end{tabular}
\caption{Cost-Accuracy Curves for Active Learning on UCI Datasets: we report accuracy after each iteration and X-axis represents cost so far in active learning. We can see AGB consistently outperforms other baselines. Results are averaged over 20 runs.}
\label{fig:uci}
\end{figure*}

We divide each dataset into initial, train and test set, consisting of 5\%, 65\% and 30\% of the data, respectively. The algorithm's performance will be better if the size of initial set is larger, but more data with ground truth is also harder to acquire. Logistic Regression is used as classifier in our experiments. We report classification accuracy on test set after each acquisition iteration. 
The results shown are averages over 20 runs.

Our main results are shown in Figure \ref{fig:uci}. We also report the average cost, query numbers and label accuracies in Tables \ref{tab:uci_cost}, \ref{tab:uci_la}, and \ref{tab:uci_query}, respectively. As we discussed in the previous section, our results show that CEAL often selects the cheapest labelers, 
and the resulting noisy annotations yield poor generalization error. ALC tends to select many expensive (and accurate) labelers and yields  the highest label accuracy amongst all methods. These  results also demonstrate that the most accurate labels may not be the most cost-effectiveness to acquire. A similar conclusion is also drawn in ~\cite{khetan2017learning,snow2008cheap}. The costs incurred by AGB and GB are relatively higher than the cost of random, but lower than the cost of ALC. This allows the AGB and GB methods to compile a larger number of instances with sufficient labeling accuracy to produce good generalization performance.  AGB and GB perform quite well in all the tasks, and AGB outperforms all other methods in all datasets, suggesting that an adaptive estimate can offer a better estimation of the expected benefits of different labelers.

\subsection{Real Dataset}
\begin{figure}[h]
    \centering
    \includegraphics[width=3.3in]{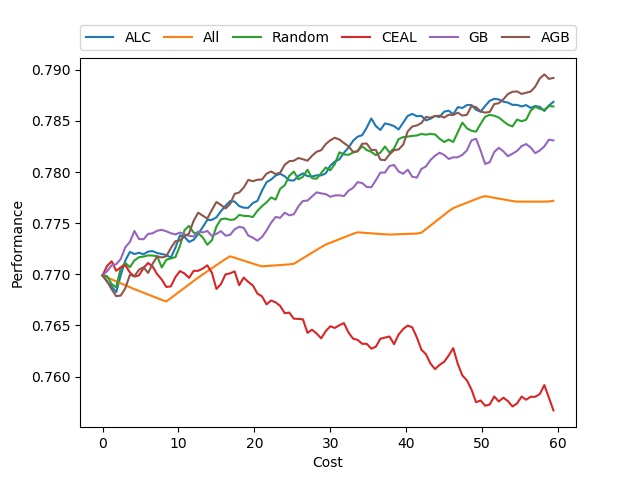}
    \caption{Cost-Effective Active Learning on Real Dataset
    , Results are averaged over 30 runs
    }
    \label{fig:real}
\end{figure}

In addition to the simulated UCI datasets, we also performed experiment using a real crowdsourcing dataset. We use a sentiment analysis dataset, introduced in ~\cite{rzhetsky2009get}, and that includes 1000 sentences labeled by five \textit{real} crowdsourcing workers. The annotators labeled each sentence along three dimensions: Focus, Polarity and Evidence. However, given the overall accuracies of the five labelers along the Polarity and Evidence labels are all very high and thus not very diverse, we only use the Focus dimension where labelers exhibit diverse accuracies. We henceforth refer to this as the Focus data set. We binarize the response variable and use bag-of-words features for training the model. After removing stopwords, the feature set consists of 292 features. The overall labeling accuracy of each real labeler is shown in Table \ref{tab:real}. We simply set a labeler's cost to be the same as the labeler's accuracy in Table \ref{tab:real}. As before, Logistic Regression is used as classifier and dataset is randomly split into 5\%, 65\% and 30\% as initial, train and test set, respectively. Result on Focus, averaged over 30 runs, are shown in Figure \ref{fig:real}.  

\begin{table}[h]
    \centering
    \begin{tabular}{c|c|c|c|c|c}
    \hline
     Labeler&  W1 & W2 & W3 & W4 & W5 \\\hline
     Label Accuracy & 0.82 & 0.931 & 0.892 & 0.904 & 0.641 \\\hline
    \end{tabular}
    \caption{Label Accuracy on Focus Dataset}
    \label{tab:real}
\end{table}

Since the price levels of Focus dataset are too close (very close to 0.9) besides W5, the performance of AGB is similar to ALC and random (Considering when all price options are same and workers' accuracies are similar, all methods will have the same effect.), but from Figure \ref{fig:real}, we can still see AGB outperforms all other methods. The result is consistent with our results on UCI datasets above.

\section{Conclusion}
In this paper, we propose that the generalization bounds from theoretical analysis for settings with noisy labels can be effectively used to address the cost-effective active learning task with labelers of varying expertise and costs. We examine the shortcomings of existing algorithms proposed for this and other similar settings, and empirically demonstrate the effectiveness of our algorithms on various datasets. It is worth noting that our proposed algorithm can also apply to choose between singly labeling and the acquisition of multiple labels per instance for majority voting strategies, which we leave for future work. However, the optimal instance-payment selection ought to account for domain, concept class, and price-accuracy tradeoffs. We use a particular generalization bound as an upper bound which allows us to account for these elements through a model-data free criterion, though clearly not optimally. We leave the design for 
a more complex model-data dependent algorithm for future work.
\bibliography{aaai}
\bibliographystyle{aaai}
\end{document}